\documentclass[letterpaper]{article} 
\usepackage{aaai2026}  
\usepackage{times}  
\usepackage{helvet}  
\usepackage{courier}  
\usepackage[hyphens]{url}  
\usepackage{graphicx} 
\usepackage{natbib}  
\usepackage{caption} 
\usepackage{algorithm}
\usepackage{algorithmic}
\usepackage{mdframed}
\usepackage{xcolor}
\usepackage{cuted}
\usepackage{fancyvrb}
\usepackage{listings}

\usepackage{newfloat}
\usepackage{listings}
\DeclareCaptionStyle{ruled}{labelfont=normalfont,labelsep=colon,strut=off} 
\floatstyle{ruled}
\newfloat{listing}{tb}{lst}{}
\floatname{listing}{Listing}
\usepackage{amsmath}
\usepackage{amssymb}
\usepackage{mathtools}
\usepackage{amsthm}
\usepackage{graphicx}
\usepackage{algorithm}
\usepackage{algorithmic}
\usepackage{tabularx}
\usepackage{booktabs}
\usepackage{mdframed}
\usepackage{multirow}
\usepackage{color}
\usepackage{xcolor}
\usepackage{colortbl}
\usepackage{xspace}

\newcommand{\model}{Math-Vision\xspace}
\newcommand{\method}{MathSE\xspace}
\newcommand{\data}{MathVL\xspace}

\newcommand{\vpara}[1]{\vspace{0.07in}\noindent\textbf{#1 }}

\title{\method: Improving Multimodal Mathematical Reasoning via Self-Evolving Iterative Reflection and Reward-Guided Fine-Tuning}
\author{
    Jinhao Chen$^{1,3}$\equalcontrib, 
    Zhen Yang$^{2,3}$\equalcontrib\textsuperscript{\rm †}, 
    Jianxin Shi$^1$, 
    Tianyu Wo$^1$, 
    Jie Tang$^2$\textsuperscript{\rm †}
}
\affiliations{
    \textsuperscript{\rm 1}School of Software, Beihang University\\
    \textsuperscript{\rm 2}Department of Computer Science and Technology, Tsinghua University\\
    \textsuperscript{\rm 3}Zhipu AI\\
    \textsuperscript{\rm †}Corresponding Authors: \texttt{yang-zhen@mail.tsinghua.edu.cn}, \texttt{jietang@mail.tsinghua.edu.cn}
}

\usepackage{bibentry}

\begin{document}

\maketitle

\begin{abstract}
Multimodal large language models (MLLMs) have demonstrated remarkable capabilities in vision-language answering tasks. Despite their strengths, these models often encounter challenges in achieving complex reasoning tasks such as mathematical problem-solving. Previous works have focused on fine-tuning on specialized mathematical datasets. However, these datasets are typically  distilled directly from teacher models, which capture only static reasoning patterns and leaving substantial gaps compared to student models. This reliance on fixed teacher-derived datasets not only restricts the model's ability to adapt to novel or more intricate questions that extend beyond the confines of the training data, but also lacks the iterative depth needed for robust generalization.
To overcome these limitations, we propose \textbf{\method}, a \textbf{Math}ematical \textbf{S}elf-\textbf{E}volving framework for MLLMs. In contrast to traditional one-shot fine-tuning paradigms, \method iteratively refines the model through cycles of inference, reflection, and reward-based feedback. Specifically, we leverage iterative fine-tuning by incorporating correct reasoning paths derived from previous-stage inference and integrating reflections from a specialized Outcome Reward Model (ORM). 
To verify the effectiveness of \method, we evaluate it on a suite of challenging benchmarks, demonstrating significant performance gains over backbone models. Notably, our experimental results on MathVL-test surpass the leading open-source multimodal mathematical reasoning model QVQ. Our code and models are available at 
\texttt{https://zheny2751\allowbreak-dotcom.github.io/\allowbreak MathSE.github.io/}.
\end{abstract}

\section{Introduction}

Multimodal large language models (MLLMs)~\cite{openai2024gpt4o,Claude3,bai2023qwen-vl,Qwen2-VL,Qwen2.5-VL,wang2023cogvlm,hong2024cogvlm2,chen2024internvl} have recently garnered significant attention for their impressive ability to integrate visual and textual information, enabling them to effectively address a variety of vision-language answering tasks~\cite{antol2015vqa,kafle2017visual,mishra2019ocr}. However, their performance tends to falter when confronted with complex reasoning challenges, such as mathematical problem-solving. In order to enhance the mathematical reasoning ability of MLLMs, existing methods~\cite{gao2023gllava,yang2024mathglmvision,cai2024geogpt4vgeometricmultimodallarge,shihu2024mathllava,zhang2024mavis,peng2024multimath,luo2025ursa} have primarily focused on fine-tuning these models on specialized mathematical datasets. These approaches typically involve distilling detailed, step-by-step reasoning from teacher models to generate rich, annotated datasets that capture mathematical problem-solving processes.

\begin{figure}[t!]
    \centering
    \includegraphics[width=\linewidth]{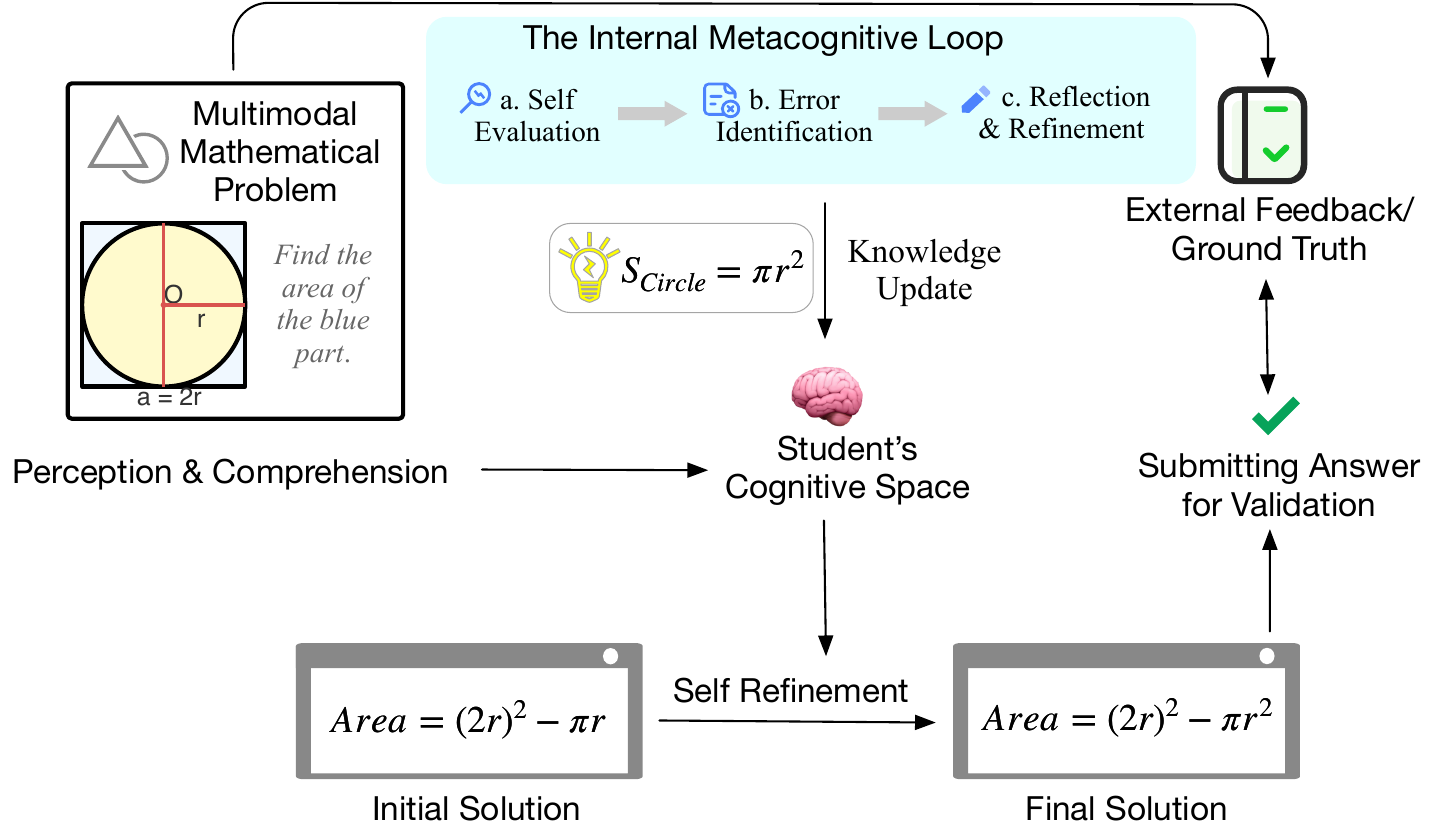}
    \caption{Illustration of the human learning process that inspires our approach. }
    \vspace{-3mm}
    \label{fig:motivation}
\end{figure}

Although previous methods have led to improvements in mathematical reasoning task, they still face notable limitations. The reliance on static, teacher-derived datasets often prevents models from adapting to novel or more intricate problems beyond the scope of the training data. Moreover, the step-by-step reasoning captured in these datasets frequently lacks the iterative depth necessary for robust generalization, leaving models ill-equipped to handle the evolving complexity inherent in mathematics. Such static and distilled datasets not only limit the dynamic, adaptive reasoning expected from student models but also widen the gap between the static patterns learned from teachers and the inherent data distribution from student models.

Inspired by human learning processes~\cite{zimmerman1990self,hattie2007power}, we recognize that effective learning process is inherently dynamic and iterative. As illustrated in Figure~\ref{fig:motivation}, human learning unfolds as a continuous cycle of instruction, practice, feedback, and improvement. In this paradigm, foundational knowledge is first acquired from teachers, distilling mathematical reasoning skills from teachers to students. This distilled knowledge forms the basis for independent practice, where students engage in self-guided problem solving to apply and reinforce what they've learned. As students tackle problems on their own, feedback plays a crucial role in identifying errors and improving their skills. Such continuous cycle of instruction, practice, feedback, and improvement enables students to progressively master mathematical problem-solving skills.

\begin{figure}[t!]
    \centering
    \includegraphics[width=0.98\linewidth]{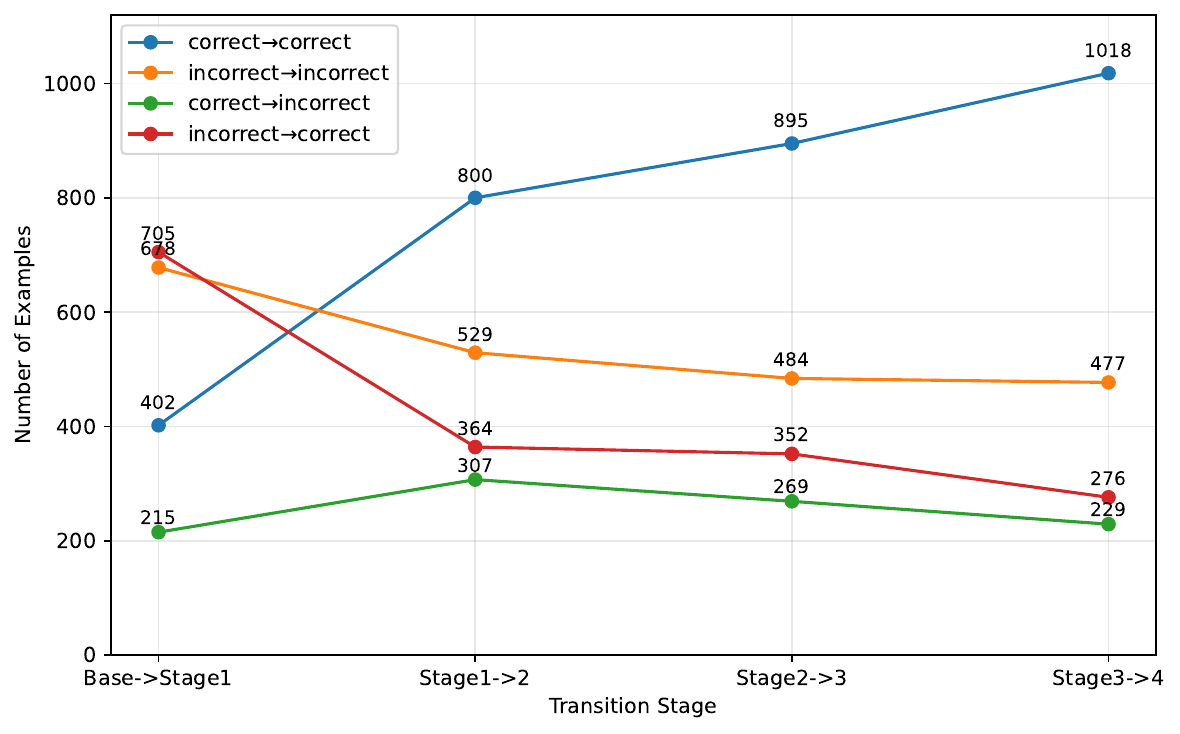}
    \caption{Accuracy changes during self-evolving process.}
    \vspace{-3mm}
    \label{fig:error_correction}
\end{figure}

Drawing on these insights, we propose a novel framework that mirrors the dynamic, iterative nature of human learning. In this work, we propose a \textbf{Mathematical Self-Evolving} method, termed as \method. \method is built on the key idea of iterative fine-tuning and feedback-driven learning, designed to continuously enhance the mathematical reasoning capabilities of MLLMs. Specifically, our approach beings by fine-tuning a base model using a subset of GPT-4o distilled Chain-of-Thought (CoT) data, enabling it to grasp foundational mathematical reasoning skills. Once this initial training phase is complete, the model is employed to generate reasoning paths on the remaining dataset. Correct reasoning paths are then identified and leveraged for further fine-tuning. This creates a self-evolving learning cycle, where the model continuously refines its reasoning abilities by learning from its previous inferences. To distinguish between correct and flawed reasoning, we introduce a specialized Outcome Reward Model (ORM), which evaluates the entire reasoning process rather than the final answers. It identifies erroneous steps and delivers detailed error analyses. By leveraging the language understanding and reasoning capabilities of large language models (LLMs), the ORM not only signals correctness but also guides the model to reflect on and learn from its mistakes.

In our iterative framework, these incorrect reasoning paths, along with ORM-provided error steps and analyses, are fed back to the GPT-4o for reflection and refinement. The resulting corrected reasoning paths form a reflection dataset that is leveraged to train the final model. Such feedback from our ORM not only enables the model to recognize and correct its mistakes but also deepens its understanding of underlying reasoning flaws. As shown in Figure~\ref{fig:error_correction}, as the self-evolving process progresses, more examples are consistently classified correctly, reflecting an improvement in overall accuracy. Such self-evolving training strategy enables the model to progressively enhance its problem-solving skills, effectively bridging the gap between static, teacher-derived datasets and the dynamic learning process characteristic of human students.

In order to verify the effectiveness and generalization of \method, we conduct experiments on three backbone models, including CogVLM2, Qwen2-VL-7B, and InternVL2.5-8B. Subsequently, we obtain a series of fine-tuned models namely \method-CogVLM2, \method-Qwen, and \method-InternVL. Experimental results demonstrate that \method achieves substantial improvements on multimodal math reasoning benchmarks, including MathVista, \data-test, MathVerse, and Math-Vision. With a parameter scale of approximately 10B, our models not only outperforms peer models of similar size but also attains performance levels comparable to state-of-the-art closed-source systems like Claude 3.5 Sonnet. Notably, the performance of our models on MathVL-test outperforms the leading open-source multimodal reasoning model QVQ~\cite{qvq-72b-preview}.

Our contributions can be summarized as follows:
\begin{itemize}
    \item \textbf{Method Perspective: }We propose a mathematical self-evolving framework (termed as MathSE) that iteratively improves multimodal math reasoning through reflection and reward-guided feedback.
    \item \textbf{Data Perspective: }We design a novel Outcome Reward Model (ORM) that provides step-wise error detection and analysis, guiding model refinement beyond mere answer evaluation.
    \item \textbf{Model Perspective: }Extensive experiments show significant performance gains on standard benchmarks, demonstrating the effectiveness of our approach. Code and model weights will be released soon.
\end{itemize}

\section{Related Work}

\subsection{Multimodal Math Reasoning}

Multimodal math reasoning requires models to process and integrate information from both textual and visual modalities to solve complex mathematical problems. 
Early approaches in this field primarily focused on text-only models with visual captions as input, such as PaLM~\cite{Chowdhery2022PaLMSL} and GPT-4~\cite{achiam2023gpt}, which demonstrated strong capabilities in language-based reasoning but struggled with visual content. 

Recent developments in multimodal large language models (MLLMs)~\cite{driess2023palme,liu2024visual,liu2024llava,wang2023cogvlm,li2022blip,dai2024instructblip,bai2023qwen-vl,chen2024internvl} have incorporated visual understanding to address these challenges. These models integrate vision encoders with language models to process images alongside text, achieving better performance in visual reasoning tasks. 

However, despite their advancements, current multimodal large language models (MLLMs) are primarily limited to answering visual question answering (VQA) tasks and performing simple reasoning~\cite{liu2024visual}. They often struggle with more complex mathematical problems that require deeper logical reasoning, precise interpretation of visual elements, or multi-step problem-solving. This limitation highlights the gap between their current capabilities and the demands of advanced multimodal math reasoning.

\subsection{Supervised Fine-Tuning and Knowledge Distillation}

Supervised Fine-Tuning (SFT) has been widely used to adapt pre-trained models to specific tasks. By leveraging labeled datasets, SFT enables models to refine their understanding of task-specific patterns and improve performance on downstream applications. In the context of multimodal math reasoning, SFT has been employed to fine-tune models on datasets that combine textual and visual mathematical problems, such as ChartQA~\cite{masry2022chartqa} and GeoQA~\cite{chen2022geoqageometricquestionanswering}. Despite its effectiveness, SFT often requires large amounts of high-quality labeled data, which can be expensive and time-consuming to obtain.

Knowledge Distillation (KD) offers an alternative approach to enhance model performance by building high-quality math QA-pairs synthesized with frontier models like GPT-4o and Claude 3.5-Sonnet. However, most distillation-based methods~\cite{alpaca,zhang2023llamaadapter,zhuang2024math,cai2024geogpt4vgeometricmultimodallarge,shihu2024mathllava} focus on scaling the dataset size or leveraging outputs from larger teacher models, often ignoring the importance of in-distribution data. This oversight limits the student model's ability to generalize beyond the teacher's capabilities. 
Our approach emphasizes the role of in-distribution data and iterative refinement, moving beyond static fine-tuning to enable continuous self-evolution.

\subsection{Reward Models for Reasoning Tasks}

Reward Models (RMs) have been extensively explored in reinforcement learning from human feedback (RLHF), where they are used to align model outputs with human preferences. Traditional Outcome Reward Models (ORMs)~\cite{cobbe2021trainingverifierssolvemath,stiennon2022learningsummarizehumanfeedback,yu2023outcomesupervised} typically assign scalar rewards based on outcome quality without considering the reasoning process. In mathematical reasoning, such models fail to provide actionable feedback for improving intermediate steps.

Recent work attempts to address this through process-based supervision~\cite{lightman2023lets,wang2023mathshepherd,luo2024improvemathematicalreasoninglanguage,wu2024exampleshighlevelautomatedreasoning}, where models are rewarded for correctness at each reasoning step. However, this paradigm requires exhaustive step-level correctness labels while potentially over-penalizing inconsequential mistakes.

In contrast, our \textbf{Outcome Reward Model (ORM)} introduces a paradigm shift by focusing on \textit{diagnostic error analysis} rather than binary step-level evaluations. Unlike process-supervised RMs that require perfect intermediate verification, our ORM identifies \textit{critical error step} in reasoning chains and provides \textit{targeted error analysis}, enabling more effective revisions based on previous reasoning paths. 

\subsection{Self-Improvement and Reflection Mechanism}

Large Language Models (LLMs) have increasingly demonstrated capabilities for self-improvement, where models leverage their own outputs to enhance performance without direct human supervision~\cite{huang2022largelanguagemodelsselfimprove,madaan2023selfrefine}. Qwen2.5-Math~\cite{yang2024qwen25mathtechnicalreport} incorporates self-improvement mechanisms into its entire pipeline, enabling it to steadily improve problem-solving accuracy, minimize errors, and enhance generalization across diverse mathematical tasks.

To further augment this self-improvement process, recent work introduces reflection mechanisms that enable models to critically analyze their own reasoning traces~\cite{lee2024reinforcementlearningreflectivefeedback,renze2024selfreflectionllmagentseffects}. ReAct~\cite{yao2023react} and Reflexion~\cite{shinn2023reflexion} integrate reasoning and reflection to allow models to reconsider incorrect outputs. These methods demonstrate the potential of iterative self-improvement but are primarily applied to text-based tasks.

In the domain of multimodal reasoning, such reflective mechanisms remain underexplored. Our framework integrates reflection with ORM-guided feedback, enabling iterative self-improvement in multimodal math reasoning tasks. By combining reflection with error-specific feedback, our model can progressively enhance its reasoning capabilities.

\section{Methodology}

In this section, we present our mathematical self-evolving framework, designed to iteratively improve reasoning capabilities through a cycle of supervised fine-tuning, reward-guided feedback, and reflection. Our method consists of three key components: (1) initial supervised fine-tuning with GPT-4o distilled data, (2) specialized Outcome Reward Model (ORM) for error detection and analysis, and (3) iterative reflection and self-improvement.

\subsection{Framework Overview}

\begin{figure*}[t]
    \centering
    \includegraphics[width=0.98\linewidth]{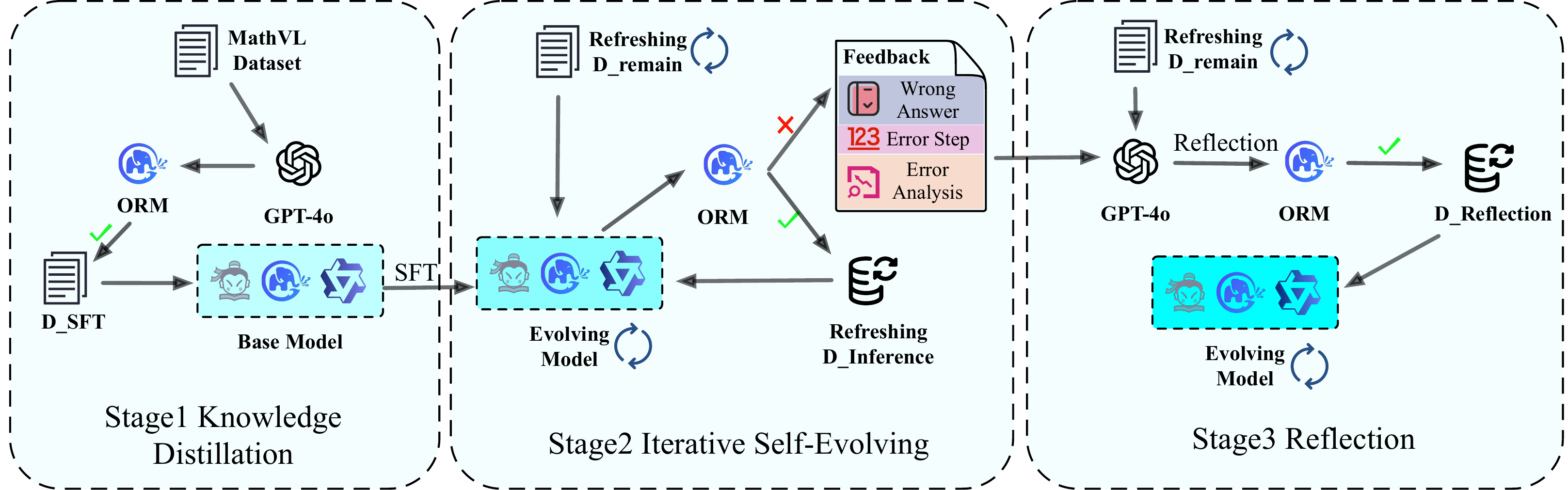}
    \caption{Overview of the \method Framework, which contains three stages to iteratively enhance mathematical reasoning abilities.}
    \vspace{-3mm}
    \label{fig:framework}
\end{figure*}

Figure~\ref{fig:framework} illustrates the overall workflow of \method framework. The process begins with fine-tuning a base model using a subset of GPT-4o distilled data. This fine-tuned model generates reasoning paths on the remaining dataset, which are then evaluated by a specialized Outcome Reward Model (ORM). The ORM identifies incorrect reasoning steps and provides a detailed error analysis. This process is repeated for multiple rounds to progressively enhance the model's reasoning ability. GPT-4o then reflects on the incorrect answer, corrects its reasoning. The improved reasoning paths are incorporated into the fine-tuning dataset to train the final model. 

\subsection{Supervised Fine-Tuning with GPT-4o Distilled Data}

Our training pipeline begins with \textbf{Supervised Fine-Tuning (SFT)} performed on the base model using high-quality curated data distilled from GPT-4o.

The fine-tuning objective is formulated as:

\begin{equation}
    \mathcal{L}_{\text{SFT}} = -\sum_{(x, y) \in \mathcal{D}_{\text{SFT}}} \log P_{\theta}(y|x)
\end{equation}

where \( \mathcal{D}_{\text{SFT}} \) denotes the distilled dataset containing GPT-4o generated reasoning paths, \( x \) represents the input query, \( y \) corresponds to the target step-by-step reasoning response, and \( P_{\theta} \) indicates the model's probability distribution.

\subsection{Specialized Outcome Reward Model (ORM)}

The specialized \textbf{Outcome Reward Model (ORM)} is a pivotal component of our framework, designed to provide both correctness evaluation and error analysis. Unlike traditional reward models that focus solely on output correctness, our ORM offers a comprehensive assessment by directly categorizing a reasoning path as either correct or incorrect. If an error is detected, the ORM pinpoints the faulty step and provides a detailed analysis of the error.

The ORM operates in the following two stages:
\begin{enumerate}
    \item \textbf{Correctness Evaluation}: The ORM evaluates the entire reasoning path \( r_i = \{s_1, s_2, \dots, s_k\} \), assigning it as either correct or incorrect.
    \item \textbf{Error Identification and Analysis}: If the reasoning path is identified as incorrect, the ORM directly specifies the step \( s_j \) where the error occurred and provides a detailed explanation \( E_i \) of the reasoning flaw that led to the mistake.
\end{enumerate}

To train our ORM, we first constructed a comprehensive training dataset consisting of 60k reasoning samples. Specifically, we collected 30k incorrect reasoning paths along with their correct solutions, and leveraged GPT-4o to automatically generate detailed annotations, including the precise location of errors and corresponding error analysis. These annotated error cases were then combined with 30k correct Chain-of-Thought (CoT) reasoning examples to form a balanced dataset. We performed Supervised Fine-tuning (SFT) on CogVLM2~\cite{wang2023cogvlm} using this curated dataset, enabling the model to effectively evaluate reasoning correctness and provide precise error analysis when necessary.

\subsection{Reflection and Self-Improvement}

Incorrect reasoning paths, along with ORM-provided error steps and analyses, are fed back into the model for reflection. We leverage the language understanding and reasoning abilities of large models like GPT-4o to prompt the model to analyze its mistakes and generate improved solutions.

The reflection process involves the following prompt format:
\begin{mdframed}
\begin{quote}
\texttt{Here is your previous solution:} \textit{[Incorrect Reasoning Path]} \\
\texttt{Error Step:} \textit{[Faulty Step]} \\
\texttt{Error Analysis:} \textit{[Explanation]} \\
\texttt{Please reflect and correct your solution.}
\end{quote}
\end{mdframed}

The refined reasoning paths \( R_{\text{reflected}} \) are combined with previously correct paths \( R_{\text{correct}} \) to form the next round of training data:
\begin{equation}
    \mathcal{D}_{\text{next}} = R_{\text{correct}} \cup R_{\text{reflected}}
\end{equation}

\subsection{Iterative Training Process}

The \method framework operates through an iterative training process that refines the model’s reasoning capabilities over multiple rounds. The process begins by generating an initial set of correct reasoning paths, \( D_{\text{SFT}} \), using GPT-4o on the entire dataset \( D \). These paths form the basis for fine-tuning the base model \( M_{\text{base}} \), yielding the initial iteration model \( M_0 \). This first step ensures that the model starts with a set of well-formed reasoning patterns.

In each subsequent round, the model generates reasoning paths on the remaining dataset \( D_{\text{remain}} \), which consists of data points not yet covered by the initial reasoning paths. These newly generated paths are then evaluated using the Output Reward Model (ORM). The ORM identifies correct paths, \( R_{\text{correct}} \), and provides feedback on incorrect paths. The incorrect reasoning paths are reflected upon and corrected by generating new paths, \( R_{\text{reflected}} \), based on the ORM feedback.

As the model progresses, the remaining dataset is updated to exclude the correct and reflected paths, and the training dataset \( \mathcal{D}_{\text{SFT}} \) is expanded by incorporating both the correct and reflected paths. This updated training dataset is used to fine-tune the model for the next round, resulting in an improved model \( M_i \). This cycle is repeated for a set number of rounds, \( T \), leading to the model’s gradual improvement.

The process can be formalized in Algorithm \ref{alg:self_evolving}, which outlines the iterative training procedure. By integrating supervised fine-tuning, reward-guided feedback, and reflection, our self-evolving framework effectively enhances multimodal math reasoning. This iterative improvement allows the model to adapt and generalize, leading to state-of-the-art performance on benchmark datasets.

\begin{algorithm}[h]
\caption{Framework of \method.}
\label{alg:self_evolving}
\begin{algorithmic}[1]
\STATE Generate correct reasoning paths \( D_{\text{SFT}} \) using GPT-4o on a subset of dataset \( D \).
\STATE Initialize base model \( M_{\text{base}} \).
\STATE Fine-tune \( M_{\text{base}} \) on \( D_{\text{SFT}} \) to obtain model \( M_0 \).
\STATE Set \( D_{\text{remain}} = D - D_{\text{SFT}} \) as the remaining dataset.
\STATE Initialize \( R_{\text{incorrect}} \) to collect incorrect reasoning paths.

\FOR{each round \( i = 1 \) to \( K \)}
    \STATE Use \( M_{i-1} \) to generate reasoning paths \( R_{\text{gen}} \) on \( D_{\text{remain}} \).
    \STATE Evaluate \( R_{\text{gen}} \) with the ORM to obtain correct paths \( R_{\text{correct}} \) and incorrect paths with feedback.
    \STATE Update \( R_{\text{incorrect}} \) with newly identified incorrect paths and their ORM feedback.
    \STATE Update remaining dataset: \( D_{\text{remain}} = D_{\text{remain}} - R_{\text{correct}} \).
    \STATE Update training data: \( D_{\text{SFT}} = D_{\text{SFT}} \cup R_{\text{correct}} \).
    \STATE Fine-tune to obtain the new model \( M_i \) using \( D_{\text{SFT}} \).
\ENDFOR

\STATE Use GPT-4o to reflect on \( R_{\text{incorrect}} \) with ORM feedback, generating reflected reasoning paths \( R_{\text{reflected}} \).
\STATE Evaluate \( R_{\text{reflected}} \) with ORM to identify correct reflected paths \( R_{\text{reflect\_correct}} \).
\STATE Update training data: \( D_{\text{SFT}} = D_{\text{SFT}} \cup R_{\text{reflect\_correct}} \).
\STATE Fine-tune to obtain the final model \( M_{\text{final}} \) using the updated \( D_{\text{SFT}} \).
\end{algorithmic}
\end{algorithm}

\section{Experiments}

In this section, we evaluate the effectiveness and generalization of our \textbf{\method} framework on multiple standard benchmarks. 

\subsection{Experimental Setup}

\subsubsection{Dataset}

Our study leverages the MathVL dataset~\cite{yang2024mathglmvision}, a comprehensive educational dataset containing 341,346 multimodal mathematics exercises specifically designed for Chinese K12 education system and several open-source dataset including GeoQA+~\cite{cao2022augmented}, Geometry3K~\cite{lu2021inter}, ChartQA~\cite{masry2022chartqa}, and UniGEO-Calculation~\cite{chen2022unigeo}. 
Furthermore, to enhance the dataset's diversity and coverage across mathematical domains, we strategically integrated carefully selected open-source datasets including MultiMath~\cite{peng2024multimath}, MAVIS\cite{zhang2024mavis}, Math-PUMA~\cite{zhuang2024math}, and MathV360K~\cite{shihu2024mathllava}. 
Our multimodal dataset spans elementary to senior high school curricula, encompassing diverse mathematical disciplines including arithmetic, algebra, geometry, probability, and applied word problems. While preserving the original question stems and multimodal contexts from MathVL, we employ \method to regenerate answer solutions. \method significantly extends the average answer length from 325 characters to 792 characters, providing more complete problem-solving procedures and more structured reasoning pathways.

\subsubsection{Public Benchmarks}

We evaluate our model on three widely used multimodal math reasoning benchmarks and our specially constructed \data-test dataset:

\begin{itemize}
    \item \textbf{MathVista}~\cite{lu2024mathvista}: A benchmark designed for visual math reasoning tasks, combining textual and diagrammatic information.
    \item \textbf{MathVerse}~\cite{zhang2024mathverse}: Covers a wide range of math problems requiring both textual and visual comprehension.
    \item \textbf{MathVision}~\cite{wang2024measuring}: Focuses on complex multimodal math problems, challenging both vision and reasoning capabilities.
    \item \textbf{\data-test}~\cite{yang2024mathglmvision}: A curated dataset designed to evaluate the integration of visual understanding and mathematical reasoning.
\end{itemize}

\subsubsection{Baselines}

We compare our model with several Multimodal Large Language Models(MLLMs), including both closed-source MLLMs~\cite{openai2023gpt4v,openai2024gpt4o,Claude3,team2023gemini} and open-source MLLMs~\cite{qvq-72b-preview,Qwen2.5-VL,glm2024chatglm,internlmxcomposer2,gao2023gllava,chen2023sharegpt4v,zhang2024mavis,peng2024multimath,luo2025ursa}.
These models represent the current state-of-the-art in multimodal reasoning and serve as strong baselines for evaluating our approach.

\subsubsection{Implementation Details}

In this study, we carefully selected three backbone models for our experiments, each chosen for their unique capabilities and performance in visual-language tasks. The models are as follows:

\begin{itemize}
\item \textbf{CogVLM2}~\cite{hong2024cogvlm2} is an open-source multi-modal large language model based on Meta-Llama-3-8B-Instruct. The model architecture supports context lengths up to 8K tokens and processes images at resolutions up to 1344 × 1344 pixels. 

\item \textbf{Qwen2-VL-7B}~\cite{Qwen2-VL} is built upon the Qwen2-7B language model backbone with a 675M ViT-based vision encoder. It features Naive Dynamic Resolution for handling arbitrary image sizes and Multimodal Rotary Position Embedding (M-ROPE) for processing textual, visual, and video positional information. 

\item \textbf{InternVL2.5-8B}~\cite{chen2024expanding} is a multimodal large language model that combines InternViT-300M-448px-V2.5 as the vision encoder with internlm2.5-7b-chat as the language model. It employs dynamic high-resolution strategies with random JPEG compression for enhanced robustness.
\end{itemize}

For reproducibility and transparency, implementation details are documented in the supplementary material, which includes SFT configuration, prompts for data generation, training data examples, and error cases.

\subsection{Main Results}

\vpara{Results on \data-test.} The experimental results on \data-test depicted in Table~\ref{tab:mathvl_results} demonstrate the effectiveness of our iterative training framework across multiple stages. 
All three models show substantial improvements, with \method-InternVL exhibiting particularly promising results by achieving the highest accuracy of 65\%.

\begin{table}[h]
\centering
\renewcommand{\arraystretch}{1.1} 
\begin{tabular}{lcc}
\toprule
\textbf{Model}            & \textbf{Training Data} & \textbf{Accuracy(\%)} \\
\midrule
\textbf{GPT-4o}           & -      & 51.05 \\
\textbf{Claude 3.5 Sonnet}& -      & 46.84 \\
\textbf{Gemini-1.5-pro}   & -      & 52.03 \\
\textbf{QVQ-72B}          & -      & \cellcolor{red!25}{52.25} \\
\textbf{Qwen2.5-VL-7B}    & -      & 50.50 \\ 
\midrule
\textbf{CogVLM2}       & -      & 30.85 \\ 
+ Distilled Data          & 100K    & 55.35 \\
+ Self-Evolving          & 240K    & 62.35 \\
+ Reflection              & 280K    & \cellcolor{blue!25}{64.70} \\
\midrule
\textbf{Qwen2-VL-7B}      & -       & 40.60 \\ 
+ Distilled Data          & 100K    & 48.80 \\
+ Self-Evolving           & 240K    & 55.15 \\
+ Reflection              & 280K    & \cellcolor{blue!25}{57.00} \\
\midrule
\textbf{InternVL2.5-8B}   & -       & 33.20 \\ 
+ Distilled Data          & 100K    & 58.45 \\ 
+ Self-Evolving           & 240K    & 64.45 \\
+ Reflection              & 280K    & \cellcolor{blue!25}{65.13} \\
\bottomrule
\end{tabular}
\caption{Accuracy on \data-test across various backbones.}
\vspace{-3mm}
\label{tab:mathvl_results}
\end{table}

\vpara{Results on Several Benchmarks.} Table~\ref{tab:public_results} presents the performance comparison between our model and baseline models across the three widely used multimodal math reasoning benchmarks: MathVista, MathVerse, and MathVision. Our approach significantly improves the performance of base models, with average score increases of 15.91\% for CogVLM2, 12.28\% for Qwen2-VL-7B, and 8.04\% for InternVL2.5-8B. Notably, \method framework shows the most significant gains on MathVista (GPS), boosting CogVLM2 by 31.06\%, Qwen2-VL-7B by 25.96\%, and InternVL2.5-8B by 15.39\%, demonstrating particularly strong capabilities on geometry-focused mathematical reasoning tasks.

\begin{table*}[ht]
\centering
\small
\renewcommand{\arraystretch}{1.1} 
\resizebox{0.98\textwidth}{!}{%
\begin{tabular}{lccccccccc}
\toprule
\textbf{Model}                 & \textbf{Method} & MathVista (GPS) & MathVista & MathVerse & MathVision & \textbf{Average} \\
\midrule
\multicolumn{7}{c}{\textbf{Closed Source Models}} \\ 
\midrule
\textbf{GPT-4V}                & - & 50.50                      & 49.90  & 50.80 & 22.76 & 43.49 \\ 
\textbf{GPT-4o}                & - & \textbf{64.71}  & 63.80  & 56.65 & 30.39 & 53.89 \\
\textbf{Claude 3 Opus}         & - & 52.91                      & 50.50  & 31.77 & 27.13 & 40.58 \\ 
\textbf{Claude 3.5 Sonnet}     & - & 64.42                      & 67.70  & 48.98 & 37.99 & \textbf{54.77} \\ 
\textbf{Gemini-1.5 Pro}        & - & 53.85                      & 63.90  & 51.08 & 19.24 & 47.02 \\
\textbf{Gemini-2.0 Flash}      & - & -  & \textbf{73.10}  & \textbf{47.80} & \textbf{41.30} & -     \\
\midrule
\multicolumn{7}{c}{\textbf{Open Source Models}} \\
\midrule
\textbf{GLM-4V-9B}             & -     & 46.12 & 46.70 & 35.66 & 15.31 & 35.95 \\
\textbf{InternLM-XC2}          & -     & 63.00 & 57.60 & 24.40 & 14.54 & 39.89 \\
\textbf{ShareGPT4V-G-7B}       & -     & 32.69 & 45.07 & 16.24 & 12.86 & 26.72 \\
\textbf{ShareGPT4V-G-13B}      & -     & 43.27 & 49.14 & 16.37 & 14.45 & 30.81 \\
\textbf{G-LLaVA-7B}            & -     & 53.40 & 28.46 & 12.70 & 12.07 & 26.66 \\
\textbf{G-LLaVA-13B}           & -     & 56.70 & 35.84 & 14.59 & 13.27 & 30.10 \\  
\textbf{Qwen2.5-VL-7B}         & -     & -     & \textbf{68.20} & 31.50 & 25.10 & -     \\  
\textbf{MAVIS-7B}              & -     & 64.10 & -     & 28.40 & -     & -     \\  
\textbf{URSA-8B}               & -     & \textbf{79.30} & 59.80 & \textbf{45.70} & \textbf{32.60} & \textbf{54.35} \\  
\textbf{Math-LLaVA-13B}        & -     & 57.70 & 46.60 & 20.10 & 15.69 & 34.51 \\  
\textbf{MultiMath-7B}          & -     & 66.80 & 50.00 & 26.90 & -     & -     \\  
\midrule
\multirow{2}{*}{\textbf{CogVLM2}}        & Base               & 39.61 & 40.85 & 25.76 & 13.20 & 29.86 \\ 
                                            & \method-CogVLM2               & 70.67 & 53.90 & 38.83 & 19.67 & 45.77 \\
\midrule                                    
\multirow{2}{*}{\textbf{Qwen2-VL-7B}}       & Base               & 43.75 & 59.40 & 35.53 & 16.92 & 37.94 \\ 
                                            & \method-Qwen               & 69.71 & 60.60 & \textbf{45.36} & \textbf{25.22} & 50.22 \\
\midrule
\multirow{2}{*}{\textbf{InternVL2.5-8B}}   & Base               & 59.13 & 58.40 & 35.66 & 17.11 & 42.58 \\ 
                                            & \method-InternVL               & \textbf{74.52} & \textbf{61.60} & 43.65 & 22.70 & \textbf{50.62} \\ 
\bottomrule
\end{tabular}}
\vspace{-2mm}
\caption{Performance on MathVista(GPS), MathVista, MathVerse, MathVision for different models.}
\vspace{-3mm}
\label{tab:public_results}
\end{table*}

\begin{table}[H]
\centering
\begin{tabular}{lcc}
\toprule
\textbf{Reflection Mechanism} & \textbf{Accuracy (\%)} \\
\midrule
w/o Reflection                         & 62.35 \\
w/ GPT-4o Feedback                     & 64.25 \\
\textbf{w/ ORM Feedback (Ours)}    & \textbf{64.70} \\
\bottomrule
\end{tabular}
\vspace{-1mm}
\caption{Comparison of different reflection mechanisms on the \data-test set.}
\vspace{-3mm}
\label{tab:reflection_ablation}
\end{table}

\subsection{Ablation Studies}

To assess the contribution of each component in our framework, we conduct ablation studies by selectively removing or modifying key modules. All experiments in this section are based on CogVLM2 as the backbone model, ensuring a consistent evaluation of the proposed methods.

\vpara{Reflection Mechanism.}
We evaluate the impact of the reflection mechanism by comparing three configurations: (1) the proposed \textbf{ORM feedback}, where the Output Reward Model generates feedback for incorrect reasoning paths; (2) \textbf{GPT-4o feedback}, where GPT-4o replaces the ORM to provide feedback; and (3) \textbf{No reflection}, where the reflection mechanism is entirely removed. The results summarized in Table \ref{tab:reflection_ablation} show that the proposed ORM feedback achieves the highest accuracy (64.70\%), outperforming both GPT-4o feedback (64.25\%) and the no-reflection baseline (62.35\%). This demonstrates the effectiveness of the ORM in providing precise feedback for improving reasoning paths, as well as the importance of the reflection mechanism in the self-evolving process.

\vpara{Distilled data vs. self-evolving training data.}
This study examines how different data generation approaches and their resulting training distributions impact final model capabilities, comparing two paradigms:
\begin{itemize}
    \item \textbf{Ours (Proposed)}: A combination of 90k GPT-4o-generated reasoning paths and 150k reasoning paths generated iteratively by our self-evolving model after filtering and refinement (90k + 150k = 240k total). This setup balances in-distribution and out-of-distribution data across different iterations.

    \item \textbf{Full GPT-4o}: A dataset of 240k reasoning paths fully generated by GPT-4o, all filtered by the Output Reward Model (ORM) to ensure correctness. This configuration lacks the iterative self-evolving mechanism, relying solely on GPT-4o-generated paths.
\end{itemize}

\begin{table}[H]
\centering
\begin{tabular}{lcc}
\toprule
\textbf{Method} & \textbf{Data Size} & \textbf{Accuracy (\%)} \\
\midrule
Base                            & -                         & 30.85          \\
Full GPT-4o                     & \(\approx 240\)K          & 58.00          \\
\textbf{Self-Evolving (Ours)}   & \(\approx 240\)K          & \textbf{62.35} \\
\bottomrule
\end{tabular}
\vspace{-1mm}
\caption{Comparison of data generation methods on the \data-test set, with improvement ratio compared to base method.}
\vspace{-3mm}
\label{tab:data_distribution_ablation}
\end{table}

The results in Table~\ref{tab:data_distribution_ablation} show that our proposed method achieves the highest accuracy (62.35\%), significantly outperforming the full GPT-4o baseline (58.00\%). While GPT-4o provides high-quality initial data, it lacks the iterative adaptability necessary to address distributional shifts effectively. In contrast, our method leverages both pre-generated and model-refined data, leading to better generalization across diverse reasoning tasks.

\vpara{Different ORM data curation.} 
To evaluate the effectiveness of our ORM design, we conducted experiments comparing two different reward modeling approaches: our proposed ORM that incorporates both error step identification and detailed error analysis, and a baseline method that only provides binary (correct/incorrect) feedback. The baseline was trained using the same training data as our ORM, but retained only the correct/incorrect judgments. We constructed ORM-2K, a balanced test set containing 1,000 correct and 1,000 incorrect reasoning paths, to assess the models' performance. As shown in Table~\ref{tab:orm_performance_comparison}, our proposed ORM significantly outperforms the binary-only baseline, achieving higher accuracy across both positive (correct) and negative (incorrect) cases. This demonstrates that incorporating fine-grained error analysis helps the model develop a more robust understanding of reasoning correctness, even when measuring only the binary judgment capability.

\subsection{Error Correction Analysis}

We analyze the effectiveness of \method in correcting its own previous errors. As shown in Figure~\ref{fig:error_correction}, the number of consistently correct samples (correct→correct) significantly increases at each transition stage—from 402 initially to 1018 by the final stage—demonstrating strong knowledge retention. Furthermore, the count of persistently incorrect examples (incorrect→incorrect) and those changing from correct to incorrect (correct→incorrect) both show a decreasing trend, reinforcing the model's improved stability in predictions. Overall, these observations highlight \method's capability to progressively enhance its accuracy by effectively leveraging past mistakes and continuously refining its learned representations.

\begin{table}[h!]
\centering
\begin{tabular}{lcccc}
\toprule
\multirow{2}{*}{\textbf{Method}} & \textbf{Data} & \textbf{Positive} & \textbf{Negative} & \textbf{Overall}  \\
                                 & \textbf{Size} & \textbf{Acc (\%)} & \textbf{Acc (\%)} & \textbf{Acc (\%)} \\
\midrule
Binary                           & 60K           & 85.40             & 99.90             & 92.65             \\
\textbf{Ours}                    & \textbf{60K}  & \textbf{94.20}    & \textbf{100.00}   & \textbf{97.10}    \\
\bottomrule
\end{tabular}
\vspace{-2mm}
\caption{Performance comparison of different reward modeling approaches on ORM-2K test set. "Binary" represents the baseline method with binary feedback, while "Ours" indicates our proposed ORM with error step identification and analysis.}
\vspace{-4mm}
\label{tab:orm_performance_comparison}
\end{table}

\section{Conclusion}

In this paper, we introduced a novel Mathematical Self-Evolving framework which significantly enhances the reasoning capabilities of MLLMs through iterative fine-tuning, reward-guided feedback, and reflection. Our approach addresses the limitations of traditional one-shot fine-tuning and traditional reward models by introducing a specialized outcome reward model and a reflection mechanism that enables the model to progressively improve its reasoning ability. Experimental results demonstrate \method's substantial performance improvements across challenging multimodal mathematical reasoning benchmarks.

\bibliography{aaai2026}

\clearpage

\section{SFT Configuration}\label{appendix: model_configuration}

Table~\ref{tab:implementation_details} presents the detailed configurations of Supervised Fine-Tuning (SFT) process. 
As shown in the table, all models share a consistent optimization strategy, adopting the AdamW optimizer with cosine decay learning rate scheduling. 
However, the total number of training steps and input resolutions vary across different architectures, reflecting the distinct design philosophies and computational characteristics of each model. 
For instance, CogVLM2 employs a higher input resolution to better capture fine-grained visual features, while Qwen2-VL-7B utilizes a dynamic resolution strategy to balance efficiency and generalization.

During SFT, all models are trained with a global batch size of 128 and a small weight decay of $5 \times 10^{-2}$ to enhance stability. 
The warmup ratio differs slightly among models, indicating that each architecture benefits from different levels of gradual learning rate ramp-up at the early stage of training.

\begin{table*}[hbpt]
    \centering
    \renewcommand{\arraystretch}{1.15}
    \resizebox{0.65\textwidth}{!}{%
    \begin{tabular}{c|ccc}  
    \toprule
     parameters    & CogVLM2 & Qwen2-VL-7B & InternVL2.5-8B  \\
    \midrule
     Total steps & 5,0000 & 88,620 & 49,000 \\
     Global Batch Size & 128 & 128 & 128   \\
     Learning Rate & $2e^{-5}$  & $1e^{-5}$ & $2e^{-5}$  \\
     Learning Rate Schedule & cosine decay  & cosine decay  & cosine decay  \\
     Warmup Ratio & 0.01 & 0.1 & 0.03  \\
     Weight Decay & $5e^{-2}$   & $5e^{-2}$   & $5e^{-2}$   \\
     Optimizer & AdamW   & AdamW  & AdamW \\
     Input Resolution  & 1344 * 1344 & Dynamic & 448 * 448  \\
    \bottomrule
    \end{tabular}} 
    \vspace{0.2cm}
    \caption{The detailed setup of the SFT procedures.}
    \label{tab:implementation_details}
\end{table*}

\section{Error Analysis}

\begin{figure}[h]
    \centering
    \includegraphics[width=1.0\linewidth]{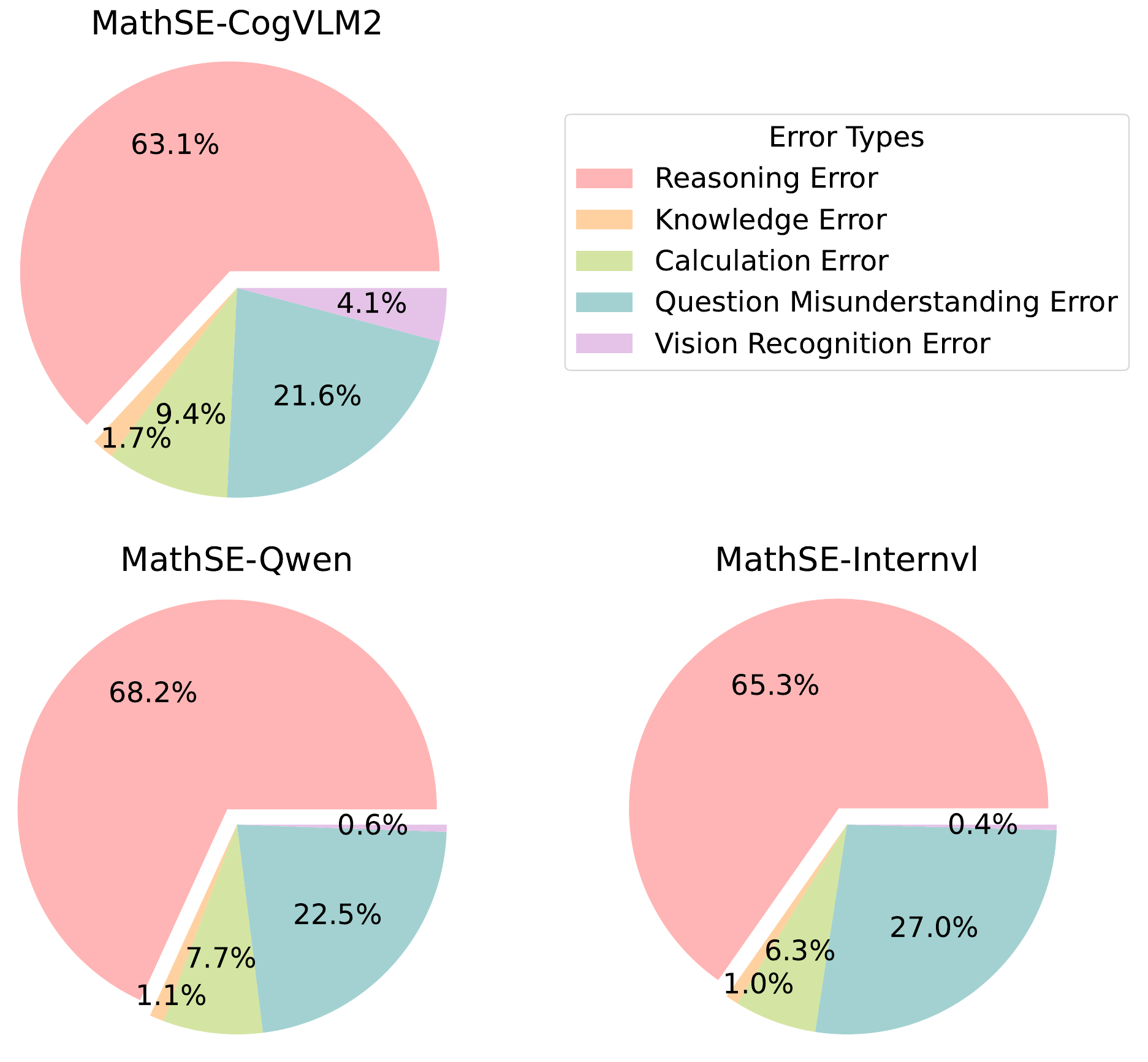}
    \vspace{-2mm}
    \caption{Distribution of different error types across three models on \data-test.}
    \vspace{-3mm}
    \label{fig:Error_Transitions}
\end{figure}

To better understand the limitations of our models, we conducted a detailed error analysis on \data-test. As shown in Figure~\ref{fig:Error_Transitions}, reasoning errors constitute the largest proportion of mistakes across all models (63.1\% for \method-Qwen, 62.9\% for \method-CogVLM2, and 65.3\% for \method-InternVL), indicating that complex reasoning remains a significant challenge in multimodal understanding. Question misunderstanding errors form the second largest category (20.8\% for \method-Qwen, 21.6\% for \method-CogVLM2, and 27.0\% for \method-InternVL), suggesting that these models sometimes struggle to properly interpret the user's intent. Knowledge errors (9.4\% for \method-CogVLM2, 7.7\% for \method-Qwen, and 6.3\% for \method-InternVL) represent the third most common error type, highlighting gaps in the factual knowledge required for solving mathematical problems. Calculation errors remain relatively infrequent across all models (1.7\% for \method-CogVLM2, 1.1\% for \method-Qwen, and 1.0\% for \method-InternVL), suggesting that basic arithmetic operations are not a major hurdle. Notably, vision recognition errors show significant variation between models, with \method-CogVLM2 exhibiting the highest rate (4.1\%) compared to \method-Qwen (0.6\%) and \method-InternVL (0.4\%), indicating potential differences in the vision processing capabilities of these models. These findings suggest that improving reasoning abilities and question understanding should be prioritized in future  mathematical MLLMs.

\section*{Prompts for Data Generation}

To generate reasoning paths and reflection feedback, we designed specific prompts for each stage. Below are some representative examples.

\definecolor{lightgray}{gray}{0.95}
\definecolor{headergray}{RGB}{120,120,120}

\newmdenv[
  backgroundcolor=lightgray,
  linecolor=white,
  skipabove=10pt,
  skipbelow=10pt,
  innerleftmargin=15pt,
  innerrightmargin=15pt,
  innertopmargin=15pt,
  innerbottommargin=15pt,
  frametitlebackgroundcolor=headergray,
  frametitlefont=\bfseries\color{white},
  frametitlerule=false
]{promptbox}

\begin{promptbox}[frametitle=Reasoning Path Generation Prompt]
You are an excellent mathematics teacher. Please provide a step-by-step detailed solution and answer to the following question based on the question and image provided. Make sure to summarize at the end by stating \texttt{"The answer to this problem is"} followed by the final result. \\
\textit{[question]} \\
\textit{[images]}
\end{promptbox}

\lstset{
  basicstyle=\ttfamily\footnotesize,
  breaklines=true,
  columns=fullflexible
}

\begin{promptbox}[frametitle=Feedback Generation Prompt]
You are a professional solution evaluation system. You need to carefully compare the predicted answer with the standard answer, evaluate its correctness, and provide improvement suggestions.

\textbf{Please evaluate the following solution:} \\
\textbf{Question:} \textit{[question]} \\
\textbf{Predicted answer:} \textit{[predict]} \\
\textbf{Standard answer:} \textit{[ground\_answer]} \\

\textbf{Please strictly follow these evaluation steps:}
\begin{enumerate}
    \item Carefully read the question requirements and evaluation criteria;
    \item Compare the predicted answer with the standard answer step by step;
    \item Check the correctness of key reasoning steps;
    \item Analyze possible causes of errors or areas that can be improved.
\end{enumerate}

\textbf{Your output must strictly follow this JSON format:}
\begin{lstlisting}
{
  "status": "CORRECT" or "WRONG",
  "error_step": "Specific step where the error occurred", 
  "error_analysis": "Analysis of the error cause",  
  "improvement_suggestion": "How to improve the answer process"
}
\end{lstlisting}
\textbf{Evaluation requirements:}
\begin{itemize}
    \item Status must be either \texttt{CORRECT} or \texttt{WRONG};
    \item If wrong, clearly indicate the step where the problem occurs;
    \item Error analysis must be specific and clearly explain the cause;
    \item If the answer is correct, still provide improvement suggestions (e.g., remove redundant reasoning, improve clarity, strengthen logic);
    \item Analysis must be objective and evidence-based, avoiding subjective guesses.
\end{itemize}

\textbf{Notes:}
\begin{itemize}
    \item If the predicted answer has a different format but identical essence, consider it correct;
    \item If multiple correct answers exist, matching any of them is valid;
    \item Do not reveal the correct final answer in the explanation.
\end{itemize}
\end{promptbox}

\begin{promptbox}[frametitle=Reflection Prompt]
\textbf{Please carefully review the following information:} \\
\textbf{Question:} \textit{[question]} \\
\textbf{Original answer (wrong\_answer):} \textit{[wrong\_answer]} \\
\textbf{Error step (wrong\_step):} \textit{[wrong\_step]} \\
\textbf{Error analysis (wrong\_analysis):} \textit{[wrong\_analysis]} \\
\textbf{And the image I uploaded}

\texttt{Please note that the authenticity of wrong\_step and wrong\_analysis may be questionable, but the original answer is confirmed to be wrong. Based on this information, please re-examine and reflect on the original incorrect answer, try to identify the errors, and regenerate a more accurate and reasonable answer. In your response, please elaborate on your thinking process and reasoning logic in detail to better verify the accuracy of the new answer. Only provide your reconsidered answer, and make sure to summarize at the end by stating "The answer to this problem is" followed by the final result.}

\end{promptbox}

\section{Training Data Examples}

\subsection{Correct Reasoning Path Example}

\begin{figure*}[h]
    \centering
    \includegraphics[width=0.98\linewidth]{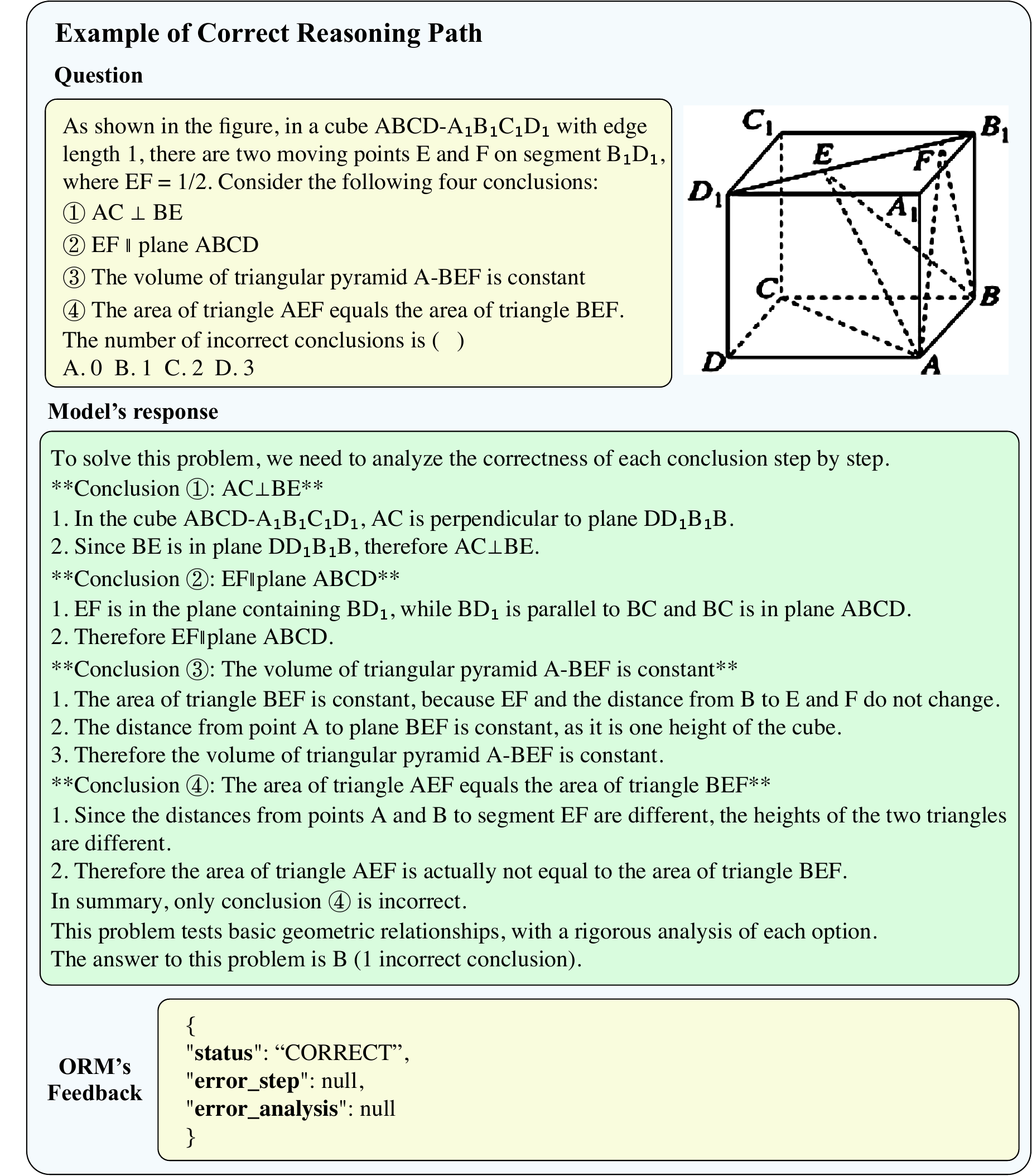}
    \caption{Example of correct reasoning path generated by \model.}
    \label{fig:Example_Correct}
\end{figure*}

Figure \ref{fig:Example_Correct} illustrates an example of correct reasoning path generated by \model. These validated reasoning paths, verified as accurate by our ORM, are subsequently incorporated into the training corpus to further enhance \model's reasoning capabilities.

\subsection{Refined Reasoning Path with ORM Feedback}

\begin{figure*}[h]
    \centering
    \includegraphics[width=0.85\linewidth]{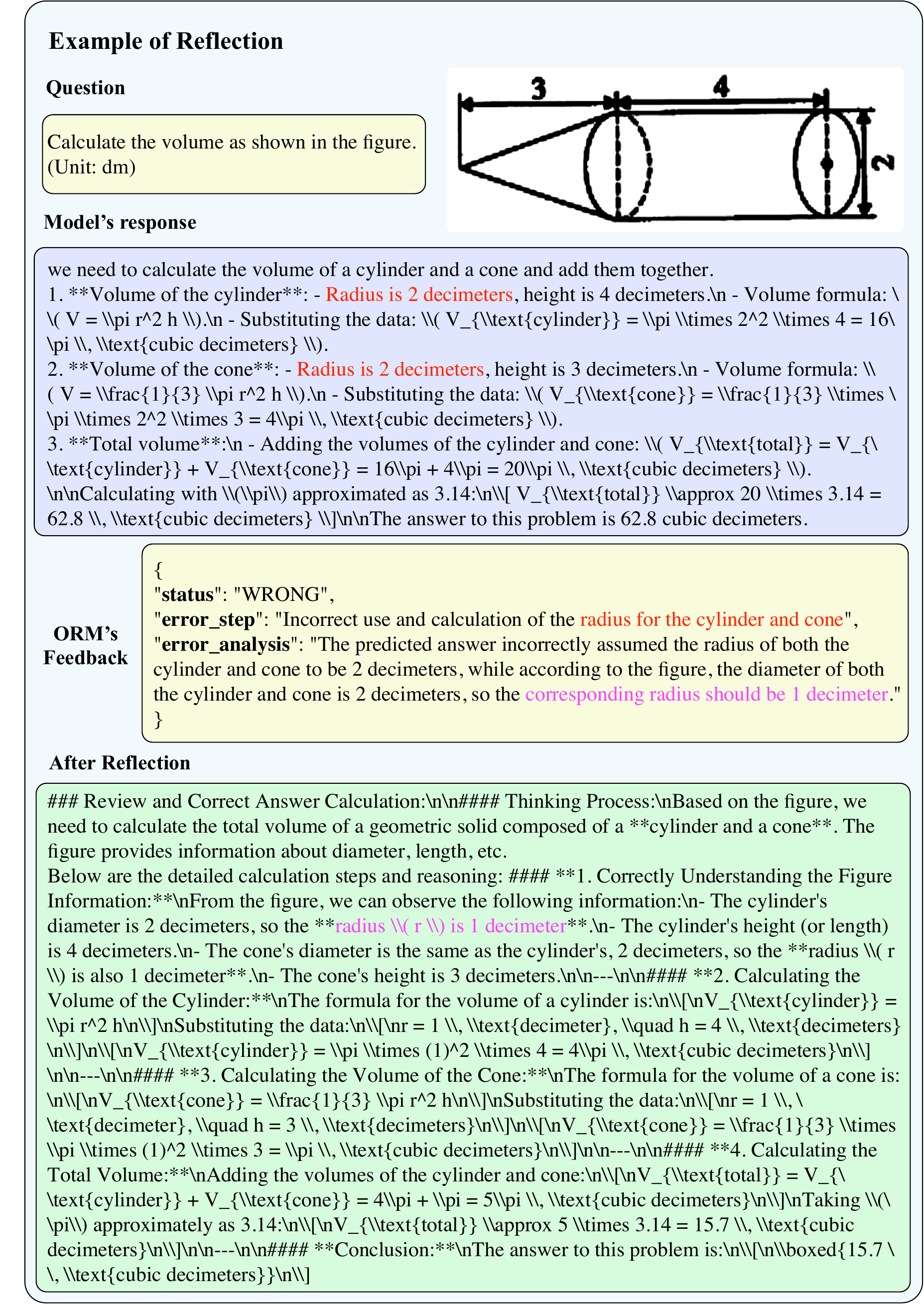}
    \caption{Example of reflection generated with incorrect reasoning path and ORM's Feedback.}
    \label{fig:Example_Reflection}
\end{figure*}

Figure \ref{fig:Example_Reflection} presents an example of incorrect reasoning with ORM feedback, demonstrating how reflection mechanisms help identify and correct errors in the reasoning process.

\section{Error Cases}

\begin{figure*}[h]
    \centering
    \includegraphics[width=0.98\linewidth]{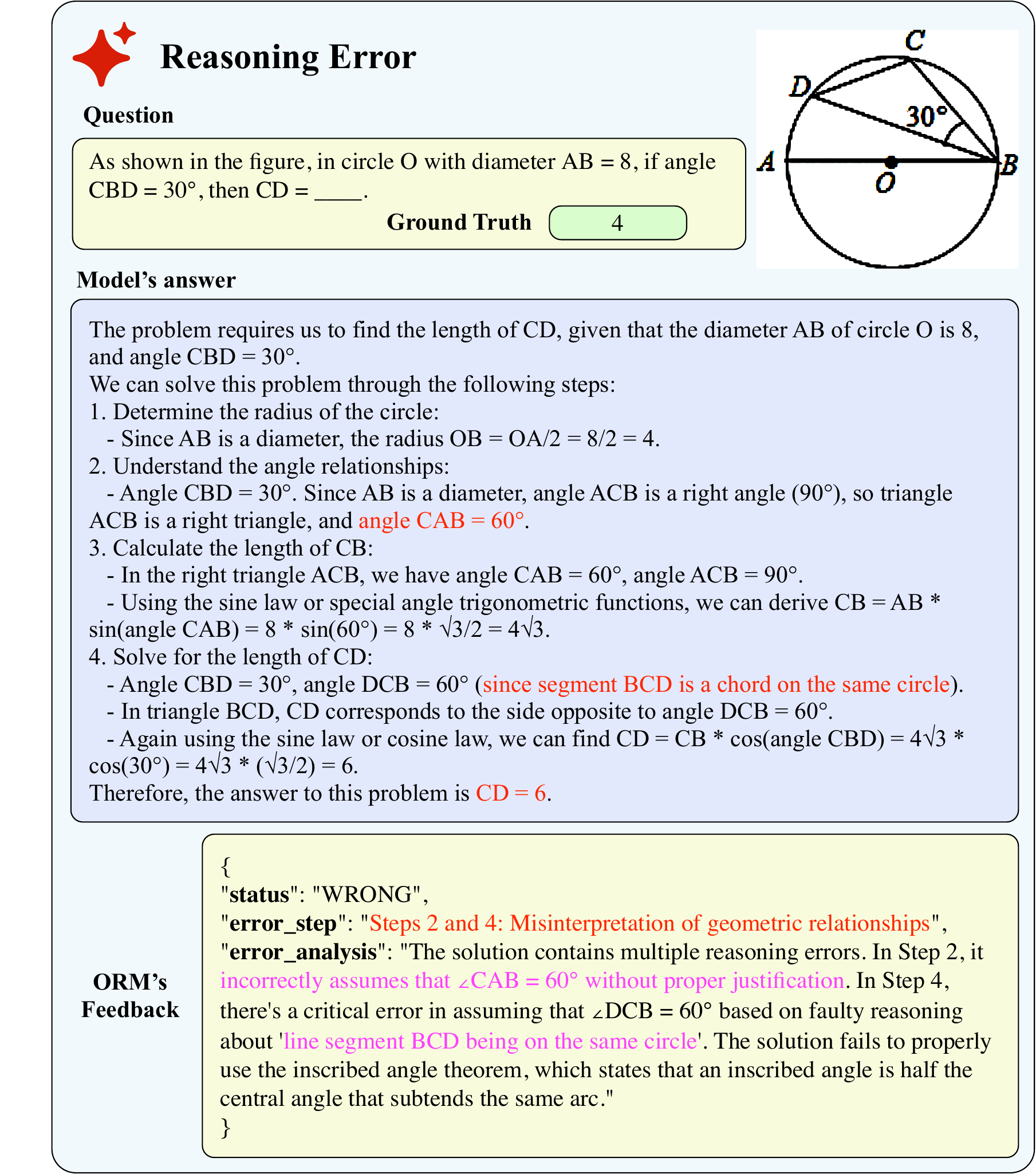}
    \caption{Example of reasoning error identified by our ORM.}
    \label{fig:Example_Error_Reasoning}
\end{figure*}

\begin{figure*}[h]
    \centering
    \includegraphics[width=0.98\linewidth]{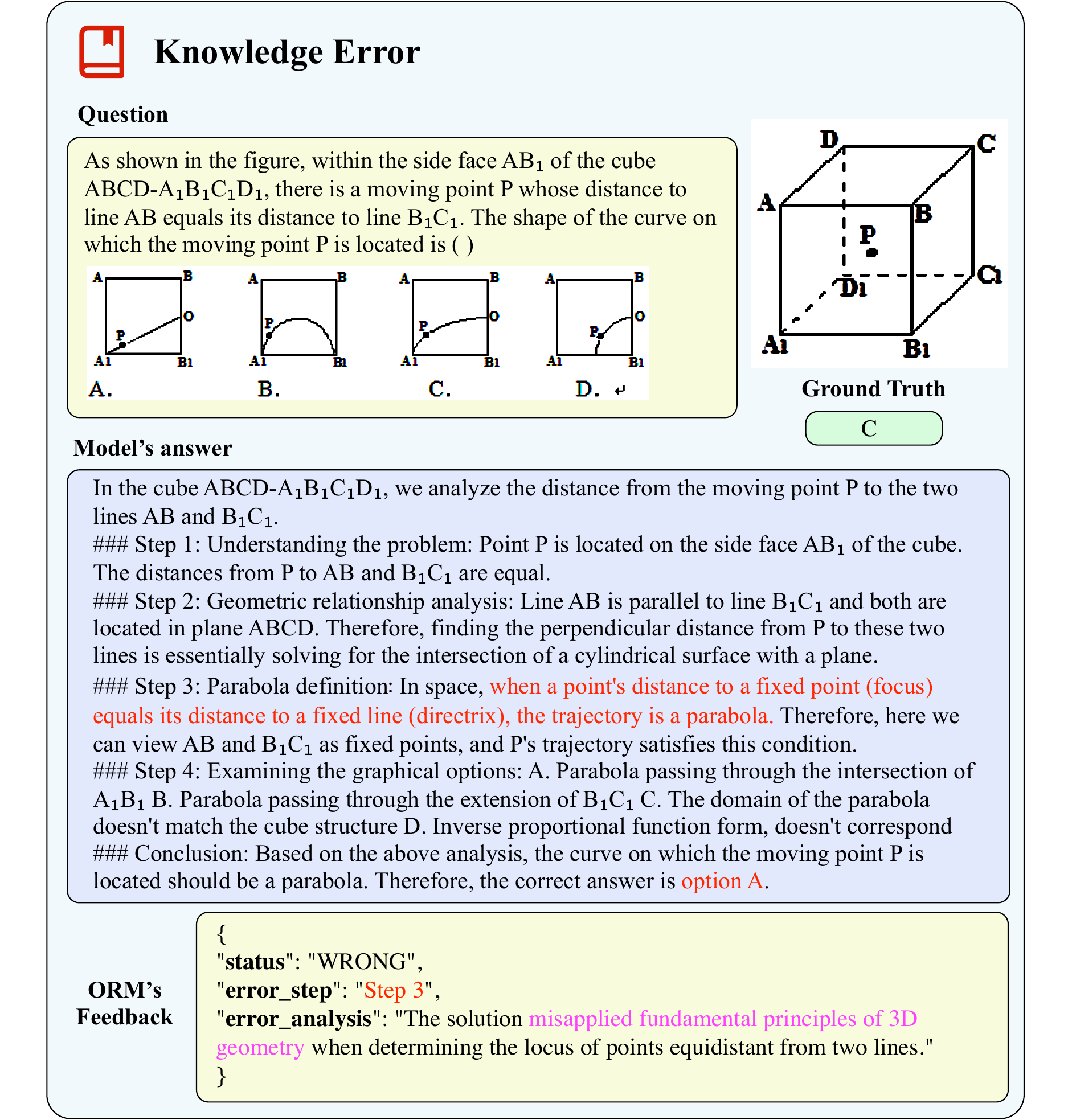}
    \caption{Example of knowledge error identified by our ORM.}
    \label{fig:Example_Error_Knowledge}
\end{figure*}

\begin{figure*}[h]
    \centering
    \includegraphics[width=0.98\linewidth]{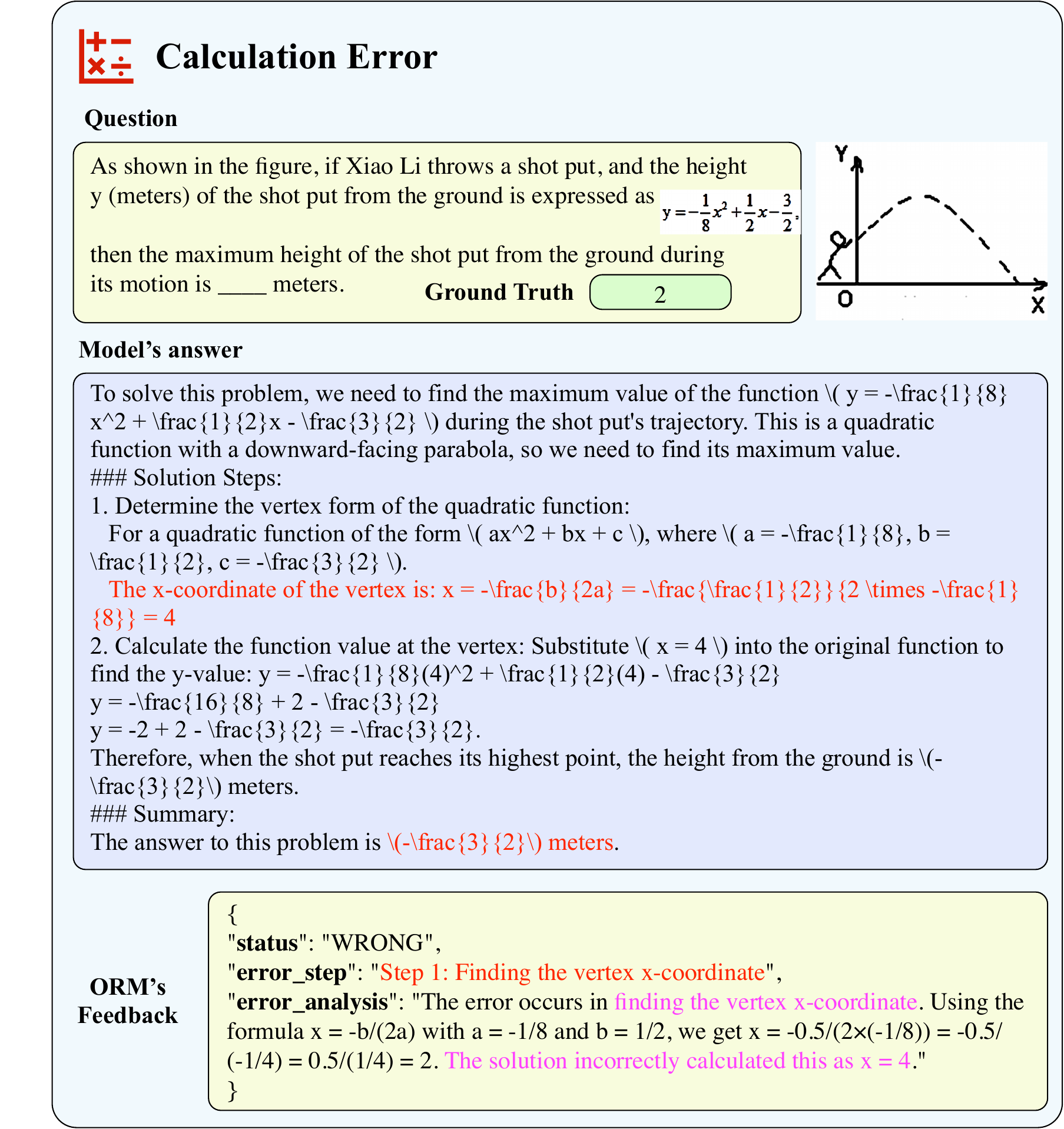}
    \caption{Example of calculation error identified by our ORM.}
    \label{fig:Example_Error_Calculation}
\end{figure*}

\begin{figure*}[h]
    \centering
    \includegraphics[width=0.98\linewidth]{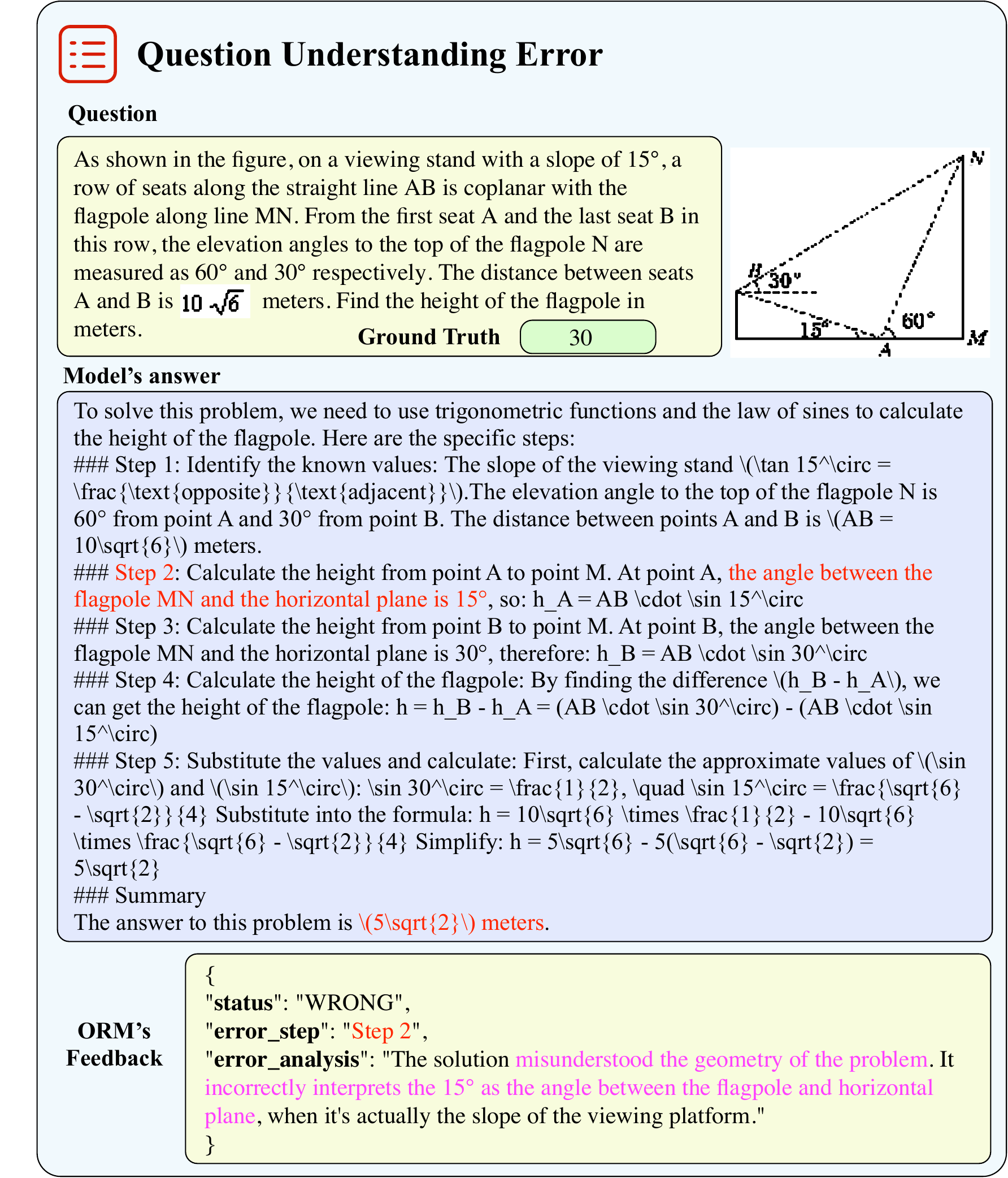}
    \caption{Example of question understanding error identified by our ORM.}
    \label{fig:Example_Error_Question}
\end{figure*}

\begin{figure*}[h]
    \centering
    \includegraphics[width=0.98\linewidth]{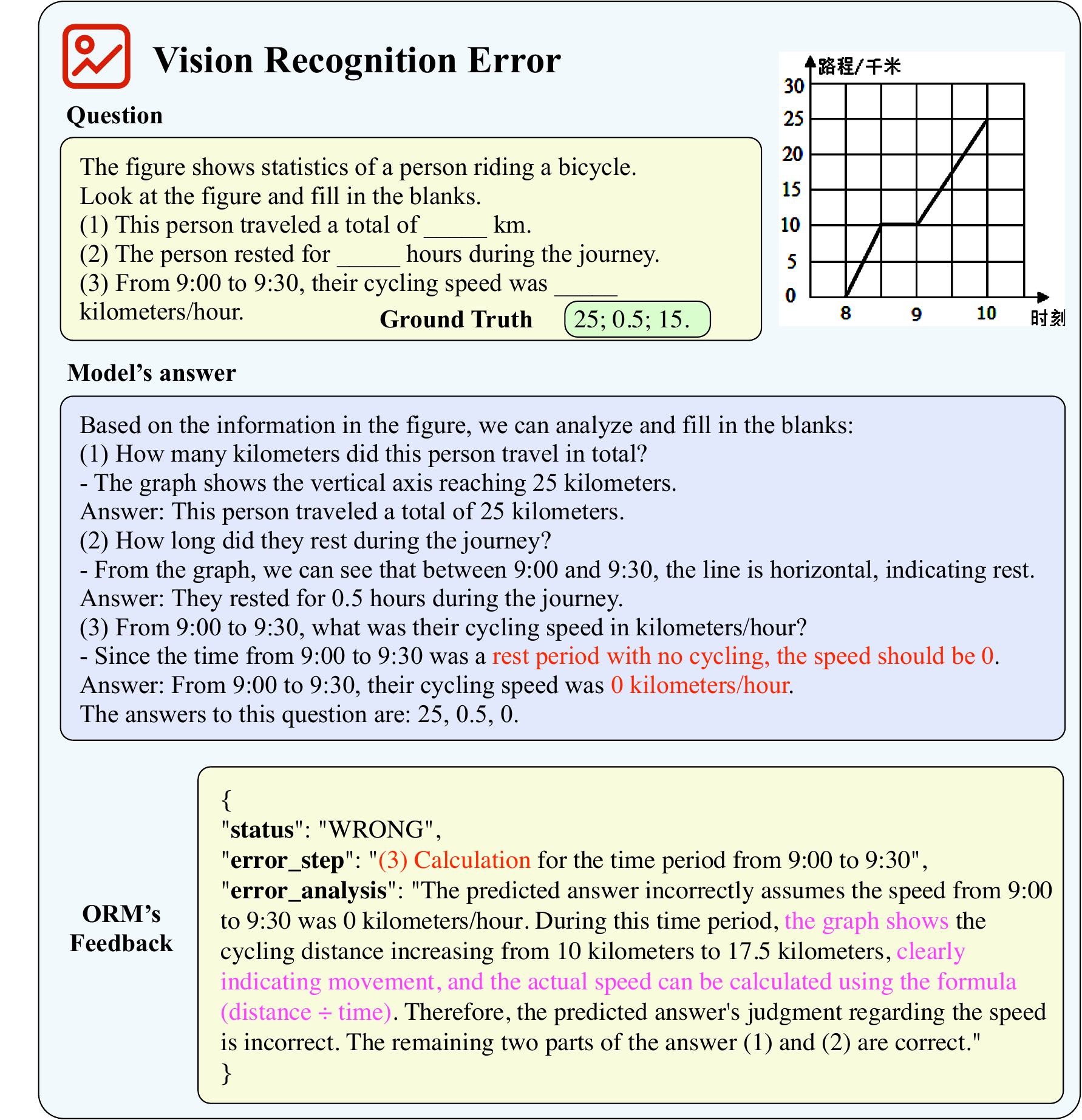}
    \caption{Example of vision recognition error identified by our ORM.}
    \label{fig:Example_Error_Visual}
\end{figure*}

We provide examples of common reasoning errors identified by the ORM, including reasoning errors (Figure~\ref{fig:Example_Error_Reasoning}), knowledge errors (Figure~\ref{fig:Example_Error_Knowledge}), calculation errors (Figure~\ref{fig:Example_Error_Calculation}), question misunderstanding errors (Figure~\ref{fig:Example_Error_Question}), and vision recognition errors (Figure~\ref{fig:Example_Error_Visual}).

\end{document}